\documentclass[10pt,twocolumn,letterpaper]{article}

\usepackage{wacv}
\usepackage{times}
\usepackage{epsfig}
\usepackage{graphicx}
\usepackage{amsmath}
\usepackage{amssymb}
\usepackage{subfigure}
\usepackage{tabularx,ragged2e,booktabs}

\DeclareMathOperator*{\argmin}{arg\,min}

\wacvfinalcopy 

\def\wacvPaperID{51} 

\ifwacvfinal\fi
\begin{document}

\title{An Epipolar Line from a Single Pixel}

\author{Tavi Halperin \hspace{2cm} Michael Werman \\
The Hebrew University of Jerusalem, Israel\\
}

\maketitle
\ifwacvfinal\thispagestyle{empty}\fi

\begin{abstract}
Computing the epipolar geometry from feature points between cameras with very different viewpoints is often error prone, as an object's appearance can vary greatly between images. For such cases, it has been shown that using motion extracted from video  can achieve much better results than using a static image. This paper extends these earlier works based on the scene dynamics.

In this paper we propose a new method to compute the epipolar geometry from a video stream, by exploiting the following observation: For a pixel $p$ in Image $A$, all pixels corresponding to $p$ in Image $B$ are on the same epipolar line. Equivalently, the image of the line going through camera $A$'s center and $p$ is an epipolar line in $B$. Therefore, when cameras $A$ and $B$ are synchronized, the momentary images of two objects projecting to the same pixel, $p$, in camera $A$ at times $t_{1}$ and $t_{2}$, lie on an epipolar line in camera $B$. Based on this observation we achieve fast and precise computation of epipolar lines.

Calibrating cameras based on  our method of finding epipolar lines  is much faster and  more robust than  previous methods.
\end{abstract}

\section{Introduction}
\let\thefootnote\relax\footnotetext{This research was supported by the Israel Science Foundation and by the  Israel Ministry of Science and Technology.}
The fundamental matrix is a basic building block of multiple view geometry and its computation is the first step in many vision tasks. This computation is usually based on pairs of corresponding points. Matching points across images is error prone, especially between cameras with very different viewpoints, and many subsets of points need to be sampled until a good solution is found. In this paper, we address the problem of robustly estimating the fundamental matrix from line correspondences in dynamic scenes.

The fundamental matrix is a $3\times 3$ homogeneous rank two matrix with seven degrees of freedom. The best-known algorithm for computing the fundamental matrix is the eight point algorithm by Longuet-Higgins \cite{longuet1981computer} which was made practical by Hartley \cite{hartley1997defense,hartley2003multiple}, and is the heart of the Gold Standard algorithm \cite{hartley2003multiple}. The overall method is based on normalizing the data, solving a set of linear equations and enforcing the rank 2 constraint \cite{luong1996fundamental}. However, it suffers from decreased accuracy when the angle between the cameras becomes wide as corresponding points become dissimilar and hard to detect, making it unsuitable for very wide angles. In such cases,  if  videos  of moving objects are available, the fundamental matrix can still be computed using motion cues.

Usually, the first step for calibrating cameras from moving objects is feature tracking, using e.g. deep features \cite{Amato2016}. Khan and Shah\cite{khan2003consistent} tracked features on a plane (people viewed from multiple surveillance cameras), and used their trajectories to compute a planar homography between the cameras. For this, they had to assume temporally synchronized cameras, long videos (minutes), and very particular movement patterns of the tracked objects. Our method shares only the first of these assumptions.

Following other papers based on motion  \cite{sinha2010camera,Calibration2016CVPR,kasten2016fundamental} (and \cite{ben2016epipolar} for still images), we use epipolar {\it lines} instead of {\it points} to compute the fundamental matrix. The use of corresponding epipolar lines instead of corresponding points has a number of good attributes; a) The exponent in RANSAC execution time depends on the size of the minimal set needed to compute the model, which is no more than 3 for epipolar lines, as opposed to 7 for points,
b) Line pairs can be filtered with motion barcodes even in very disparate views where points cannot. Indeed, three corresponding pairs of epipolar lines are enough to compute the fundamental matrix \cite{hartley2003multiple}. The epipolar lines in each image intersect at the epipole, and the one-dimensional homography between the lines can be recovered by the 3 line correspondences. The 3 degrees of freedom for the 1D homography, together with the 4 degrees of freedom of the epipoles, yield the 7 parameters needed to compute the matrix.

Sinha and Pollefeys \cite{sinha2010camera} used the silhouette of a single moving object to find corresponding epipolar lines to calibrate a network of cameras. Ben-Artzi et al. \cite{Calibration2016CVPR} accelerated Sinha's method using a similarity measure for epipolar lines based  on motion barcodes defined in \cite{ben2015event,pundik2010video}. This line motion barcode was also used in \cite{kasten2016fundamental} to find corresponding epipolar lines and is the most relevant paper to ours. In that paper, they found corresponding epipolar lines by matching all pairs of lines between the images using the motion barcode. We propose to accelerate this process by drastically reducing the search space for matching epipolar lines, utilizing pixels which record multiple depths.

On top of that, we use centroids of detected foreground areas as a proxy for an object's location, following Meingast et al. \cite{meingast2007automatic} who used it as features for correspondences of tracks from a multi-target tracking algorithm. Theoretically, estimating geometric properties based on fuzzy measurements such as areas resulting from foreground segmentation, or their centroids could be error prone. However, as shown by \cite{meingast2007automatic}, and again by our experiments, this method is robust and accurate, and when followed by a global optimization step its accuracy can be further increased. We propose such a step to refine the epipole, by better approximating the intersection of the epipolar lines. To this end, we develop a general efficient algorithm that, given a set of lines, finds a point with the minimal sum of distances to all the lines ($L1$ metric), in addition to the usual sum of squared distances ($L2$ metric).

As in previous methods, we assume stationary cameras and that moving objects have been extracted by background subtraction.

The contributions of this paper are: i) A novel algorithm to calibrate a pair of synchronized video cameras, based on motion barcodes with much smaller complexity compared to the state-of-the-art, while maintaining robustness and accuracy; ii) An epipole refinement procedure, which leads to more accurate camera calibration.

\section{Motion Barcodes}

Motion barcodes of lines are used in the case of synchronized stationary cameras recording a scene with moving objects. Following background subtraction we have a binary video, where ``0" represents static background and ``1" moving objects.

\begin{figure}
\centering
\includegraphics[width=1\linewidth]{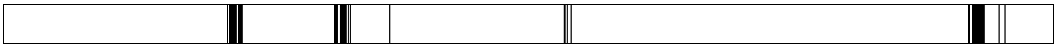}			
\caption{
A motion barcode $b$ of a line $l$ is a vector in $\{0,1\}^N$. The value of $b_l(i)$ is ``1" when a moving object intersects the line in frame $i$ ({\it black entries}) and ``0'' otherwise ({\it white entries}).
\label{fig:point_barcode}}
\end{figure}

Given such a video of $N$ binary frames, the motion barcode of a given image line $l$ is a binary vector $b_l$ in $\{0,1\}^N$ where $b_l(i)=1$ iff a moving object intersects  $l$ in the $i^{th}$ frame, \cite{Calibration2016CVPR}. An example of a motion barcode is shown in Figure ~\ref{fig:point_barcode}.

\begin{figure}
\centering
\includegraphics[width=0.85\linewidth]{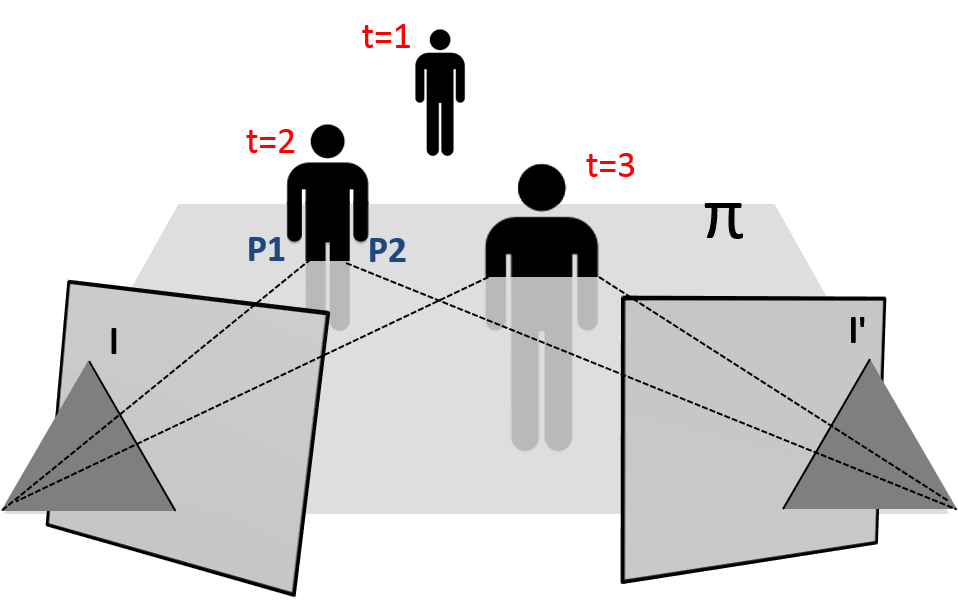}
\caption{Illustration of a scene with a moving object viewed by two video cameras. The lines $l$ and $l'$ are corresponding epipolar lines, and $\pi$ is the 3D epipolar plane that projects to $l$ and $l'$. At time $t=1$ the object does not intersect the plane $\pi$, and thus does not intersect $l$ or $l'$ in the video.  At times $t=2,3$ the object  intersects the plane $\pi$, so the projections of this object on the cameras intersect the epipolar lines $l$ and $l'$. The motion barcode of both $l$ and $l'$ is the same: $(0,1,1)$ }
\label{fig:line_barcode}
\end{figure}

The case of a moving object seen by two cameras is illustrated in Figure ~\ref{fig:line_barcode}. If the object intersects the epipolar plane $\pi$ at frame $i$, and does not intersect the plane $\pi$ at frame $j$, both motion barcodes of lines $l$ and $l'$ will be $1,0$ at frames $i,j$ respectively. Corresponding epipolar lines therefore have highly correlated motion barcodes, and the similarity measure between motion barcodes $b$ and $b'$ is their normalized cross correlation 
\cite{ben2015event}.

\section{Epipolar Lines}
Corresponding epipolar lines are projections of epipolar planes, 3d planes that go through both camera centers. Pixels are projections of 3d rays through a camera center. 

The search for corresponding epipolar lines in this paper is based on finding two different corresponding pixels in camera $B$ to a given pixel in camera $A$. These two correspondences are necessarily on an epipolar line in the other camera. The cue for matching, if there is no auxiliary information, such as tracking, color, or reliable shape features, is co-temporal movement.

The following notation is used throughout the paper: \\
\begin{tabular}{ll}
 $p,q,r$ & pixels \\
 $p_A^{t}$ & pixel $p$ imaged in camera $A$ at time $t$ \\
 $\overleftrightarrow{q_B \quad r_B}$ 
 & line between pixels $q$ and $r$ in camera $B$ \\
\end{tabular}
\\

Given a pixel $p_{A}$ imaged at times $t$ and $s$, the corresponding pixels in Image $B$, $q_B^t$ and $r_B^s$ are on the epipolar line, $\overleftrightarrow{q_B \quad r_B}$.
Likewise, $p_A$ is a point on the epipolar line in image $A$ corresponding to the epipolar line $\overleftrightarrow{q_B \quad r_B}$ in image $B$.

The algorithm for finding pairs of corresponding epipolar lines has two main steps, (i) finding two different pixels in $B$ corresponding to a single pixel $p$ in $A$, which results in a single epipolar line in $B$ and (ii) finding the corresponding epipolar line in $A$, chosen from the pencil of lines through $p$, which gives a corresponding pair of epipolar lines. 

\begin{figure}[t]
    \centering
    \subfigure[]{\includegraphics[width=0.9\columnwidth]{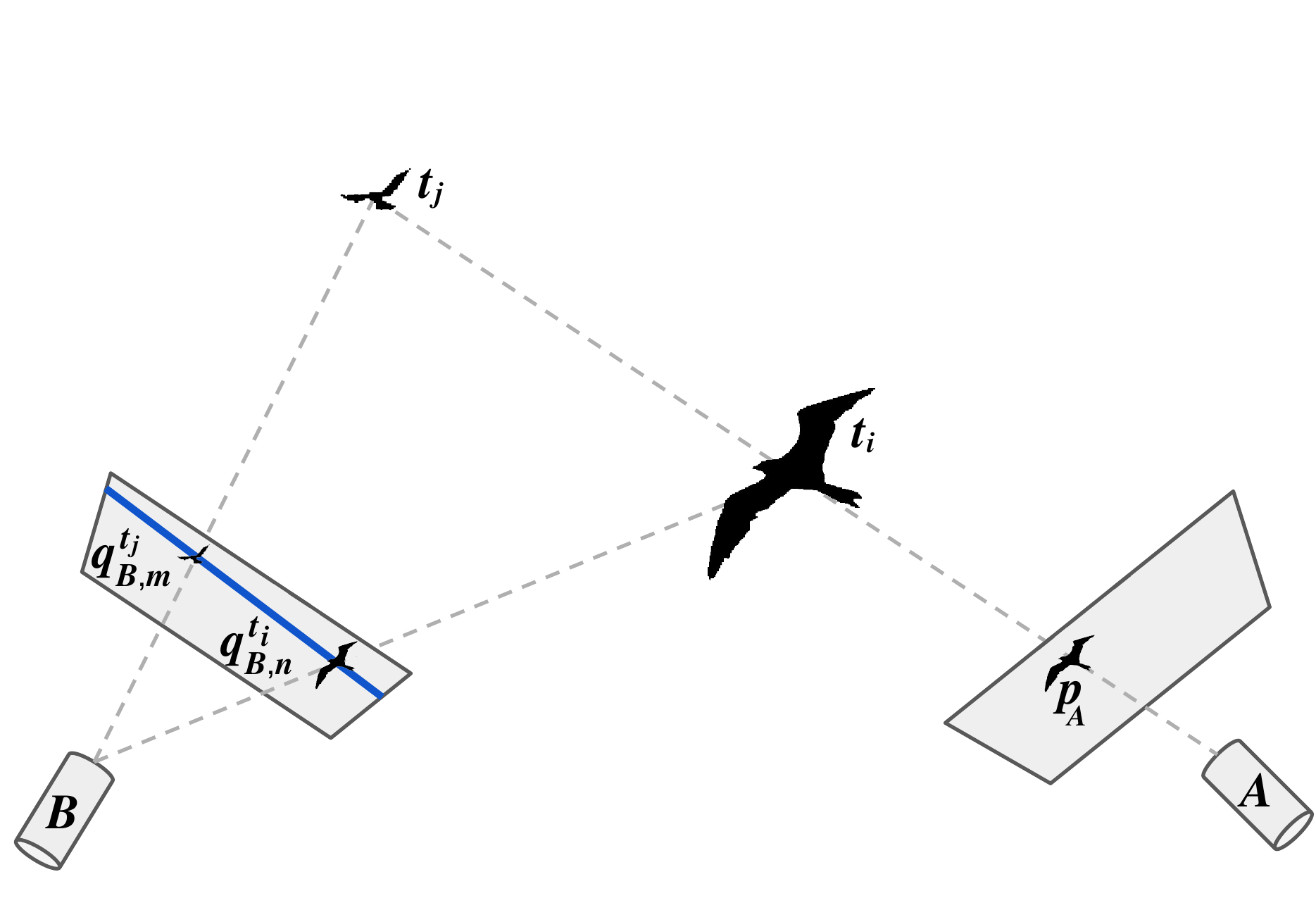}} \\
    \subfigure[]{\includegraphics[width=0.9\columnwidth]{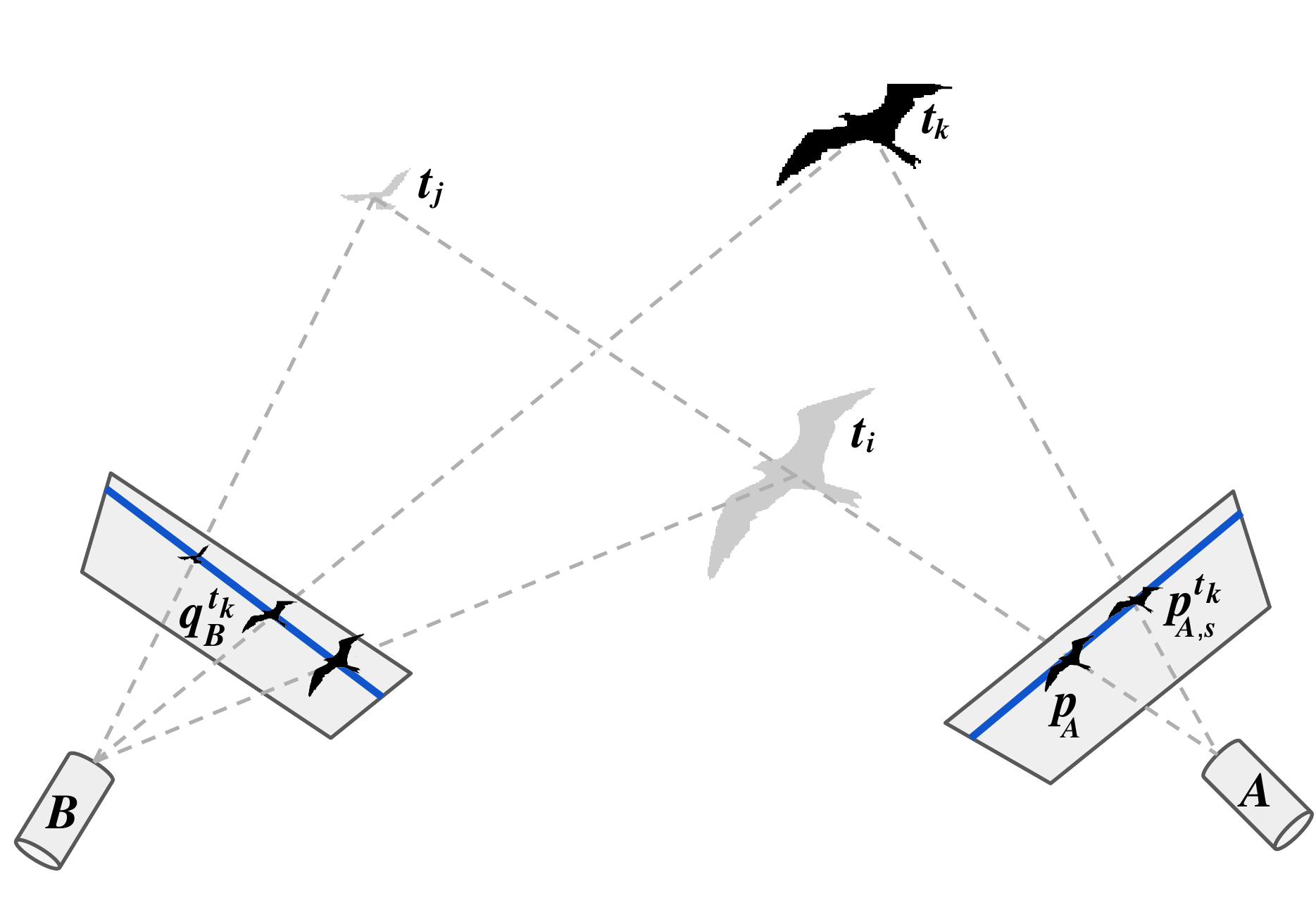}} \\

 \caption{Basic building blocks of our algorithm. 
 (a) Two pixel correspondences of a single pixel lie on an epipolar line which is the projection of the ray from camera $A$ through $p_A$.
 (b) A third pixel $q^{t_k}_{B}$ is found on the line. A matching pixel $p^{t_k}$ is chosen from camera $A$ by exploiting the similarity of motion barcodes between matching epipolar lines.}
\label{fig:bblock_alg}
\end{figure}

\begin{figure}[t]
    \centering
{\includegraphics[width=0.9\columnwidth]{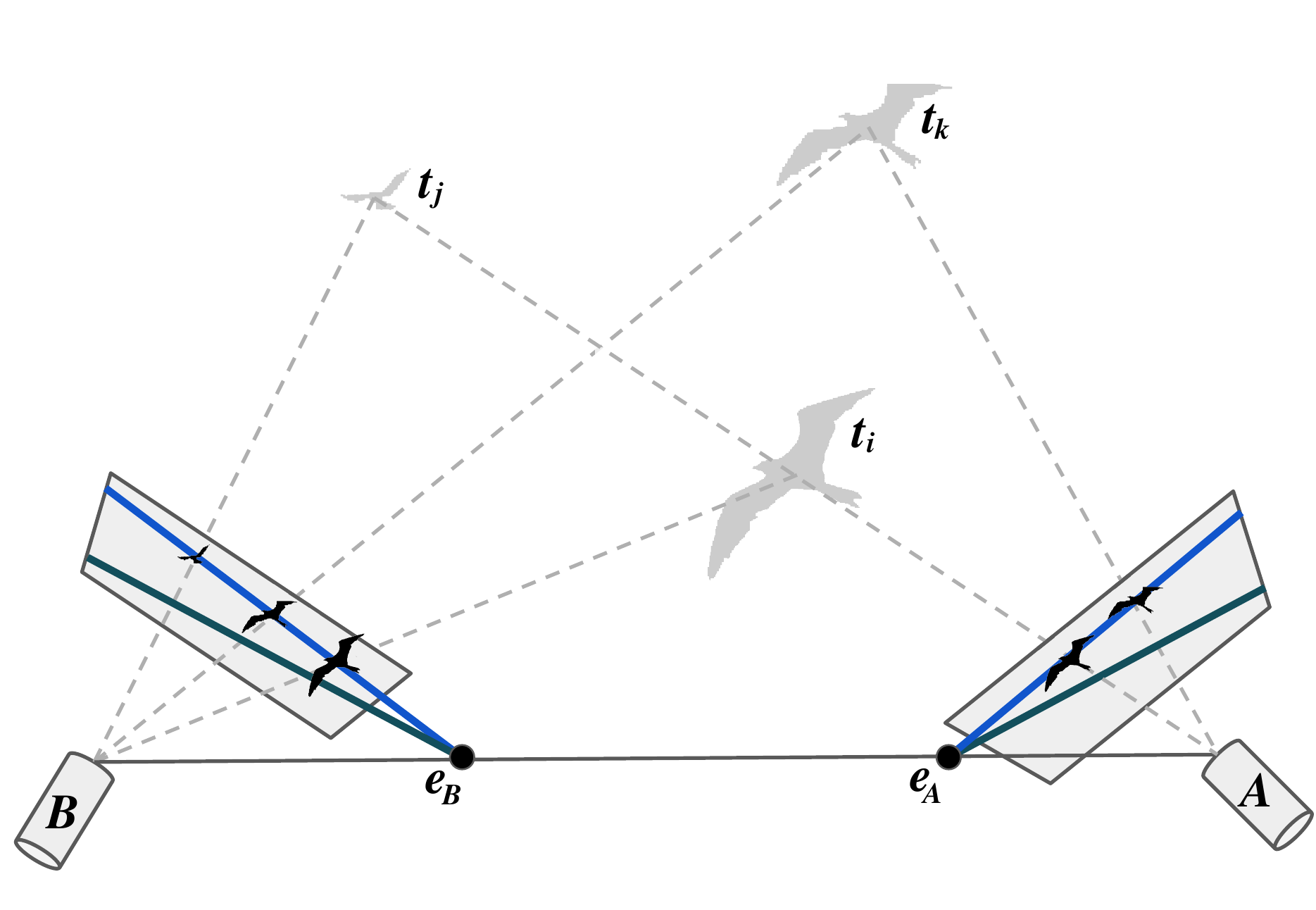}} \\
 \caption{Recovering epipoles from two pairs of epipolar lines.}
\label{fig:epipole_recovery}
\end{figure}

\section{Algorithm}
Our algorithm assumes background subtraction, and the only pixels we are interested in are the centers of mass of the detected moving objects. Therefore, for the rest of this paper, whenever we refer to a pixel $p$ it is assumed to be the center of mass of some moving object.

\subsection{Point to Line}
For two frames in camera $A$ taken at times $t_i,t_j$, and both containing the same pixel $p_{A}$, we will look at all the pixels from these frames in camera $B$, that is: $\{q^{t_i}_{B,1},q^{t_i}_{B,2} \dots \}$ and $\{q^{t_j}_{B,1},q^{t_j}_{B,2} \dots \}$. Each ordered pair of pixels $(q^{t_i}_{B,m}, q^{t_j}_{B,n})$ from the two frames gives an epipolar line candidate, and let $\Lambda_{B}=\{ \overleftrightarrow{q^{t_i}_{B,m} \quad q^{t_j}_{B,n}}\}$ be the set of all such lines, see Figure \ref{fig:bblock_alg}(a).

We now turn to find a matching line in $A$ for each candidate line in $\Lambda_{B}$. For each $l_B \in \Lambda_B$, we attempt to find a third frame at time $t_k \neq t_i,t_j$ containing an additional pixel $q^{t_k}_B$ on $l_B$. Such points usually exist in real videos. Now for such a $t_k$, let $\{p^{t_k}_{A,1},p^{t_k}_{A,2} \dots \}$ be all the pixels in camera $A$ at time $t_k$, and let $\Lambda_A$ be the lines in camera $A$ between $p_{A}$ and every such pixel, that is: $\Lambda_{A}=\{\overleftrightarrow{p_{A} \quad p^{t_k}_{A,s}} \}$, Figure \ref{fig:bblock_alg}(b). Finally, the line $l_{A} \in \Lambda_{A}$  whose motion barcode has the highest normalized cross-correlation to our $l_B$'s barcode is chosen as $l_B$'s partner, and partners with normalized cross-correlation above a certain threshold are considered a pair of possible corresponding epipolar lines. In order to proceed to the next stage we need at least two pairs of candidate lines.

It is worth noting that although the correspondence relation between each pair of line candidates is itself symmetric with respect to the roles of cameras $A$ and $B$, the process producing these pairs is not; reversing the roles of cameras $A$ and $B$ and running the same algorithm as above, may result in a different set of pairs of line candidates.

When there is enough motion in the scene, using pixels in $A$ that have 3 or more corresponding pixels in $B$ produces much better matches, and with far fewer false positives, since we can easily check if all these correspondences in $B$ are indeed co-linear. In such cases our algorithm runs much faster and is more robust; first, because we have fewer line candidates to check, and second, the chances of coincidentally having 3 pixels on the same line in 3 given frames are very low, thereby reducing the probability of errors.

\subsection{Third Line}
\label{sec:third_line}
We use RANSAC to estimate the location of the epipoles. We sample two pairs of putative corresponding epipolar lines from the previous step, $l_{A,1},l_{B,1}$ and $l_{A,2},l_{B,2}$, with the probability to sample a pair proportional to its matching score (normalized cross-correlation). The intersection of the pairs $l_{A,1},l_{A,2}$ and $l_{B,1},l_{B,2}$ suggests two epipoles locations, $e_A$ and $e_B$ respectively (Figure ~\ref{fig:epipole_recovery}). In order to compute the 1D line homography, a third pair of lines is required. If such a pair is available we skip the following step and move directly to the validation step. Otherwise, we pick a random frame $s$ and connect all foreground objects to the epipoles with lines $T_A=\{\overleftrightarrow{p_A^{s} \quad e_A}\}$, and $T_B=\{\overleftrightarrow{q_B^{s} \quad e_B}\}$. The third correspondence is found by matching the barcodes of lines from $T_A$ with those of $T_B$ and taking the best candidate to be the third pair.

These three pairs determine the 1D line homography, which together with the epipoles is sufficient to compute the fundamental matrix. 

\subsection{Validation} 
\label{validation}
The validation step is carried out for each RANSAC iteration, to evaluate the quality of the estimated epipoles and homography. Similarly to \cite{ben2016epipolar}, we compute the 1D line homography between the 3 pairs of lines, sample uniformly 10 lines from the pencil around $e_A$, transform them to the pencil around $e_B$, and compute the barcode cross-correlation between the 10 pairs of lines. The epipoles and homography with the highest score are used to compute the fundamental matrix between the cameras. 

\begin{figure}[t]
    \centering
{\includegraphics[width=0.9\columnwidth]{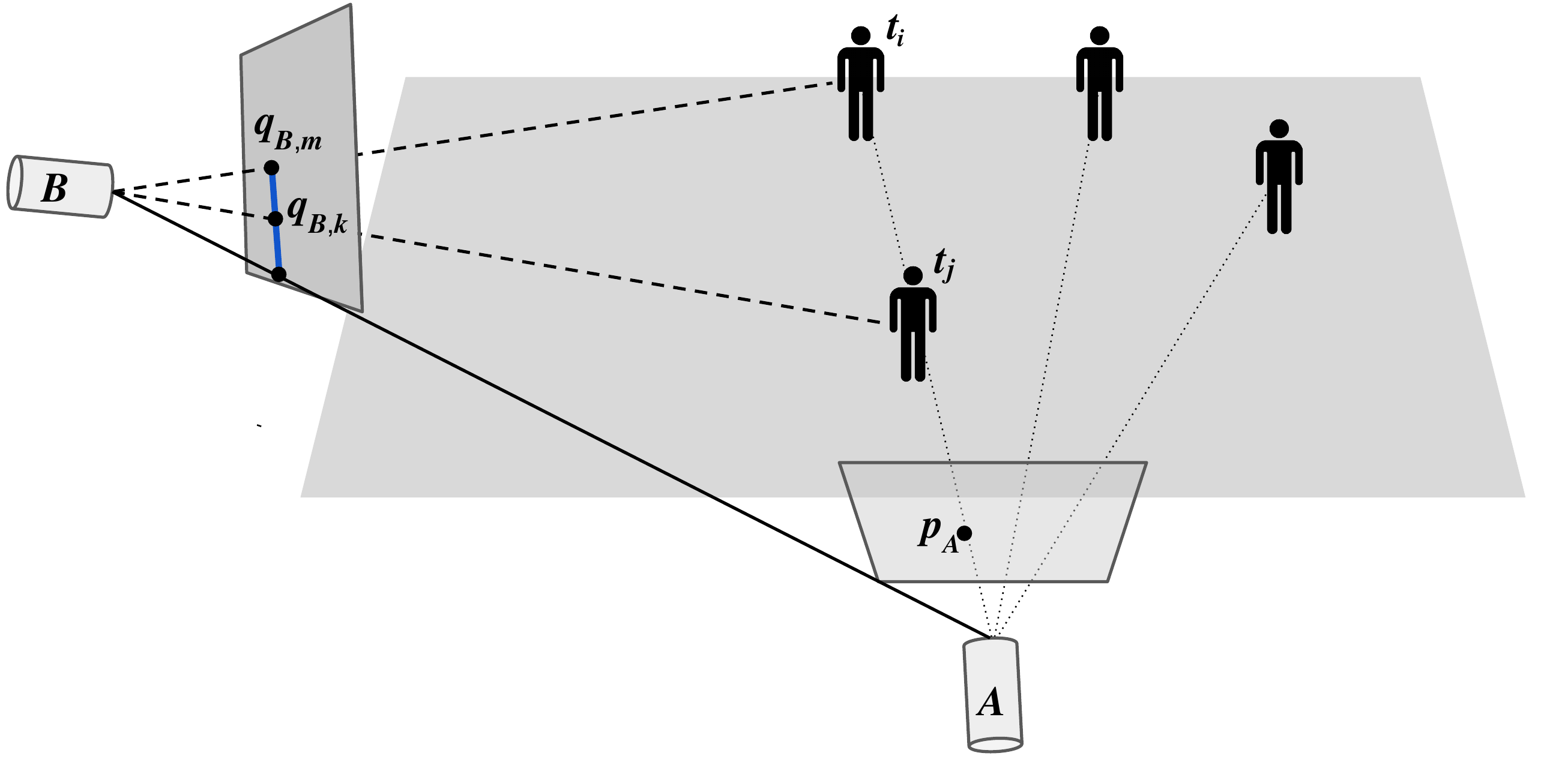}} \\
 \caption{
Scheme of planar motion. There is no other pixel imaged on the epipolar line in $A$ containing the pixel $p_A$. To match the pixel to the corresponding epipolar line $\protect\overleftrightarrow{r_{B} \quad q_{B}}$ in image $B$ we compute NCC between a \textit{point barcode} around $p$ and the \textit{line barcode} of $\protect\overleftrightarrow{r_{B} \quad q_{B}}$.}
\label{fig:planar}
\end{figure}

\section{Extensions}
\subsection{Planar motion}
\label{sec:planar_motion}
Our algorithm does not work on pure planar motion due to the major requirement of two distinctive objects having different depths on a single ray from the camera. However, in the special configuration of one camera on the plane and the other off it, the location of the epipole in the off-plane camera frame may be recovered. In this variant of the algorithm, We compute candidate lines $\Lambda_{p}$ in $B$ for a point $p_A$, where the on-plane camera plays the role of camera $A$. We exploit the following facts, (i) there is no motion outside the plane, and (ii) camera $B$ is off the plane. As a consequence, all the motion visible on the epipolar line through $p$ is concentrated around $p$ (see Figure \ref{fig:planar}). We then sample $p$'s barcode from a disc around it, instead of sampling from a line, and use NCC with the barcodes of $\Lambda_p$. The one with highest score is kept. We only recover epipolar lines in $B$, which is not enough to run the validation step, to choose the correct epipole among all intersections of lines. Instead, we ignore all lines whose matching score with their corresponding point falls under a certain threshold, and vote for the epipole by maximal consensus voting among the remaining lines. This step is carried out using RANSAC, where two lines are drawn in every iteration, their intersection yields the candidate epipole, and the number of lines which agree with the epipole is counted. The candidate with the maximal set of inliers is chosen as the epipole. A common definition for inliers in this scenario is the one introduced by \cite{kasten2016fundamental}, which we adopt for our controlled experiments. That being said, when the epipole is inside the image boundaries a simpler approach which works well is measuring whether the perpendicular distance between the epipole and a line is below a certain threshold.
As a side effect, this process allows camera $A$ to be wide-angle with extreme lens distortion, a useful attribute for such a scenario (see the scene coverage of the planar camera in Figure \ref{fig:square_closeup}). We do not measure image lines in $A$, thus lens distortion does not affect the computations, because from the point of view of camera $A$ we are only interested in rays through pixels, and those are not altered by lens distortion. This side effect may even benefit accuracy, which is improved by incorporating the increased amount of epipolar lines in $B$ imaged in $A$, and possibly larger angles between pairs of lines.

\subsection{Static objects}
In certain cases, features of multiple objects projected to the same point may be extracted. A dynamic object can occlude a static one, for which a different kind of feature (e.g. SIFT\cite{lowe2004distinctive}) can be detected, or the scene can even be fully static with multiple objects detected at the same image point, such as semi transparent surfaces. Various algorithms exist to separate reflections from transmitted light (for example \cite{levin2007user,li2014single,shih2015reflection}). Two features extracted from the separated layers at the same location, with their  matched corresponding points on different locations in the other camera, will produce an epipolar line. See example in Figure \ref{fig:glass}.

\subsection{Coupling with other features}
In addition to motion barcodes other types of features can guide the matching process. For example, two objects imaged on $p$ having certain colors (identifiable from other viewing points) will constrain the search in $B$ for objects with matching colors. More complex features such as deep features could be used, for example in a natural scene with a large number of moving objects, we can isolate one kind of moving object, e.g. butterflies, and process only their locations.

\section{Epipole Refinement}
We refine the estimated epipoles and epipolar geometry using inlier lines. Line pairs are defined as inliers if they agree with the epipoles via the measure Kasten et. al introduced in \cite{kasten2016fundamental}. They measure the area between a given line and a true epipolar line intersecting it at the central vertical line of the image. We use the same threshold of 3 times the width of the frame, as the threshold below which the line is an inlier. For the refinement process we do not have the true epipoles, but use instead our estimated ones from the validation process. Note that the inlier percentage in the controlled experiments in Tables \ref{tb:cubethin_results} and \ref{tb:cube_results} is computed using the \textit{ground truth} epipoles, in order to get the true number of inliers. 
\paragraph*{Point-line distance} 
The perpendicular (signed) distance between a line $l=(l^1,l^2,l^3)$ and a point $p=(p^1,p^2,p^3)$ can be expressed as a dot product $l\cdot p$, where $l$ is normalized such that $\Vert (l^1,l^2)\Vert=1$, and $p$ is normalized such that $p^3=1$ \cite{faux1979computational}.
\paragraph*{$L2$ refinement}
This is simply least-squares, used to compute a refined epipole $e_{L2}$ minimizing

\begin{equation}
e_{L2}=\argmin\limits_e\sum\limits_{i}{(l_{i}\cdot e)^2}
\end{equation}
given the inlier lines $\{l_1,\dots \}$. The complexity is linear in the number of lines.
\paragraph*{$L1$ refinement}
The $L1$ loss
\begin{equation}
loss_{L1}(e) = \sum\limits_i{\vert l_{i}\cdot e\vert}
\end{equation}
is a convex function, and is linear inside each of the cells of the line arrangement, hence its global minimum $e_{L1}$ is obtained on an intersection of two of the lines, which is a vertex of a cell. In general, in the presence of outliers, $L1$ refinement is more robust than $L2$.
\paragraph*{Iterative $L1$ minimization}
The complexity of using Brute Force to find $e_{L1}$ is $O(n^3)$ where $n$ is the number of lines, since there are $O(n^2)$ intersections, requiring $O(n)$ calculations each. We propose an efficient iterative algorithm to find $e_{L1}$ in $O(n^2)$. First, we compute the arrangement of the lines, a process that can be done by topological sweeping in $O(n^2)$ \cite{edelsbrunner1986topologically}. Next, we pick an intersection $q$ at random, and split the lines to two sets $neg$ and $pos$, composed of the lines with $l_{i}\cdot q<0$ and $l_{i}\cdot q\geq 0$, respectively. We get
\begin{equation}
loss_{L1}(q) = \sum\limits_{l_i\in pos}l_i\cdot q-\sum_{l_i\in neg}l_{i}\cdot q = (\sum_{l_i\in pos}l_{i}-\sum_{l_i\in neg}l_{i})\cdot q
\end{equation}
Moving to a neighbor $r$ of $q$ might cause a line $l_j$ to switch between $pos$ and $neg$, as the line is incident to $q$ but not to $r$, i.e. $l_j\cdot q=0$ and $l_j\cdot r\neq 0$. As a consequence, after moving $l_j$ from $pos$ to $neg$ (or vice versa) if necessary, and updating the sums, we can efficiently compute $loss_{L1}(r)$. At each step, updating the sum requires changing the sums by adding/subtracting at most one line, thus taking $O(1)$.
We examine the neighbors of $q$, and move to the one with lowest loss. We traverse the arrangement until the minimal point is found, visiting at most $n^2$ intersections, where at each we spend $O(1)$ computations.

\paragraph*{Fundamental matrix refinement}
Our refinement algorithm works as follows; (i) Compute refined epipole locations based on inlier pairs of lines. Recall that since we do not know the true epipoles we rely on our estimated epipoles to identify inlier lines.
(ii) Perform RANSAC iterations similar to the 'third line' step (section \ref{sec:third_line}), but in which we sample all three frames at every iteration. In each of the frames we connect lines from the pixels to the refined epipoles, and take the best matching pair of lines. (iii) The three best pairs (one pair of lines for each frame) are sufficient to compute the epipolar geometry, which gets a score using the validation process described in section \ref{validation}. To overcome errors introduced by outlier lines, we compare the validation score of the initial fundamental matrix with those computed using $L2$ refinement and $L1$ refinement. The final epipoles and homography are those with the highest validation score.

\def\tabletext{\small}
\newcommand{\specialcell}[2][c]{%
  \begin{tabular}[#1]{@{}c@{}}#2\end{tabular}}
\newcommand{\cmark}{\ding{51}}%
\newcommand{\xmark}{\ding{55}}%

\renewcommand{\tabcolsep}{0.1cm}
\begin{table*}[t]
  \centering
  \begin{tabular}{lcccccccc}
  \toprule[1.5pt]

\specialcell{\bf \tabletext} & 
\specialcell{\bf \tabletext Running} & 
\specialcell{\bf \tabletext Average Number} & 
\specialcell{\bf \tabletext Percent} & 
\specialcell{\bf \tabletext Mean} & 
\specialcell{\bf \tabletext With $L2$} & 
\specialcell{\bf \tabletext $L1{+}L2$}
\\

\specialcell{\bf \tabletext} & 
\specialcell{\bf \tabletext Time} & 
\specialcell{\bf \tabletext of Barcodes} & 
\specialcell{\bf \tabletext of Inliers} & 
\specialcell{\bf \tabletext Error} & 
\specialcell{\bf \tabletext Refinement} & 
\specialcell{\bf \tabletext Refinement}
\\ \midrule
  
\tabletext Kasten et. al \cite{kasten2016fundamental} & 
\specialcell{\tabletext $583.6$ sec.} & 
\specialcell{\tabletext $36928$} &
\specialcell{\tabletext $67.8\%$} & 
\specialcell{\tabletext $0.79$}	& 
\specialcell{\tabletext --} &  
\specialcell{\tabletext --} &  	
\\

\tabletext Ours & 
\specialcell{\tabletext $2.5$ sec.} &
\specialcell{\tabletext $480.1$}	& 
\specialcell{\tabletext $31.1\%$} & 
\specialcell{\tabletext $0.83$} & 
\specialcell{\tabletext $0.78$} & 
\specialcell{\tabletext \textbf{0.76}} & 
\\

\bottomrule[1.5pt] \\
\end{tabular}

\caption{
\label{tb:cubethin_results}
Results on the \textit{thin cubes} dataset. Mean symmetric epipolar distance was calculated over 21 camera pairs.
}
\end{table*}

\renewcommand{\tabcolsep}{0.1cm}
\begin{table*}[t]
  \centering
  \begin{tabular}{lcccccccc}
  \toprule[1.5pt]

\specialcell{\bf \tabletext} & 
\specialcell{\bf \tabletext Running} & 
\specialcell{\bf \tabletext Average Number} & 
\specialcell{\bf \tabletext Percent} & 
\specialcell{\bf \tabletext Mean} & 
\specialcell{\bf \tabletext With $L2$} & 
\specialcell{\bf \tabletext $L1{+}L2$}
\\

\specialcell{\bf \tabletext} & 
\specialcell{\bf \tabletext Time} & 
\specialcell{\bf \tabletext of Barcodes} & 
\specialcell{\bf \tabletext of Inliers} & 
\specialcell{\bf \tabletext Error} & 
\specialcell{\bf \tabletext Refinement} & 
\specialcell{\bf \tabletext Refinement}
\\ \midrule
  
\tabletext Kasten et. al \cite{kasten2016fundamental} & 
\specialcell{\tabletext $557.6$ sec.} & 
\specialcell{\tabletext $36928$} &
\specialcell{\tabletext $71.67\%$} & 
\specialcell{\tabletext $0.31$}	& 
\specialcell{\tabletext --} &  
\specialcell{\tabletext --} &  	
\\

\tabletext Ours & 
\specialcell{\tabletext $9.8$ sec.} &
\specialcell{\tabletext $1894.9$}	& 
\specialcell{\tabletext $31.7\%$} & 
\specialcell{\tabletext $0.31$} & 
\specialcell{\tabletext $0.31$} & 
\specialcell{\tabletext \textbf{0.30}} & 
\\

\bottomrule[1.5pt] \\
\end{tabular}

\caption{
\label{tb:cube_results}
Results on the \textit{cubes} dataset, comprised of 10 camera pairs.
}
\end{table*}

\section{Experiments}
\label{sec:experiments}
We  evaluated our algorithm on real and simulated video streams. Since this approach is novel, there are no existing suitable real datasets with ground truth calibration.

The authors of \cite{kasten2016fundamental} provided us with their synthetic datasets \textit{cubes} and \textit{thin cubes}, comprised of 5 and 7 cameras, respectively. We adopted their area measure and used the same threshold for defining inliers.

Our algorithm shares with theirs the RANSAC procedure, whereas the acceleration in our algorithm stems mostly from the first step in which we find putative corresponding epipolar line pairs. We reimplemented their method with our barcode sampling and matching, to allow a fair comparison. We tested our method on all camera pairs from both datasets (see Tables \ref{tb:cubethin_results} and \ref{tb:cube_results} for quantitative comparison with state of the art). Kasten et al. \cite{kasten2016fundamental} sampled a constant number of line barcodes, spaced equally on the image boundaries. On average, their number of barcodes is 20x-75x the amount of barcodes we sample. The tables also show the percent of inliers among the line candidates, which are all candidates in our method, and the 1000 line pairs with top matching score in \cite{kasten2016fundamental}. Our average Symmetric Epipolar Distance (SED) is comparable to \cite{kasten2016fundamental}, and even outperforms their method when applying global refinement, while reducing the time complexity by two orders of magnitude. The run time increase caused by $L2$ refinement is negligible, and the increase due to $L1$ refinement depends on the actual number of lines taken into calculation, but it never took more than a second to optimize.

The code was written in Python, and all experiments were conducted on a standard desktop computer with no GPU acceleration.

\subsection{Real videos}
To validate our method on real video examples, we captured several scenes with various types of motion.  

\begin{figure*}[t]
    \begin{tabular}{cc}
{\includegraphics[width=1\columnwidth,keepaspectratio] {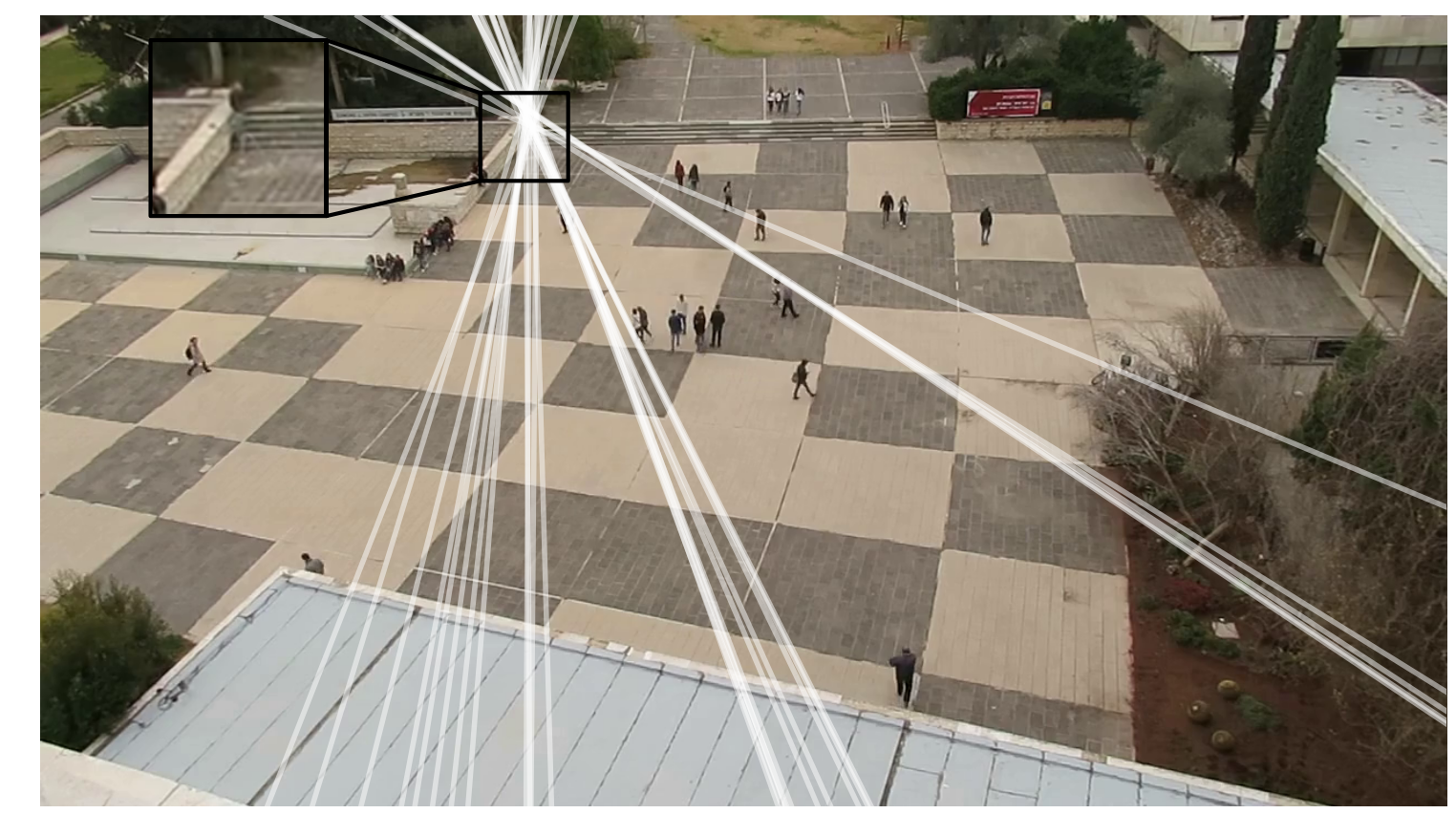}} 
&
{\includegraphics[width=1\columnwidth,keepaspectratio]{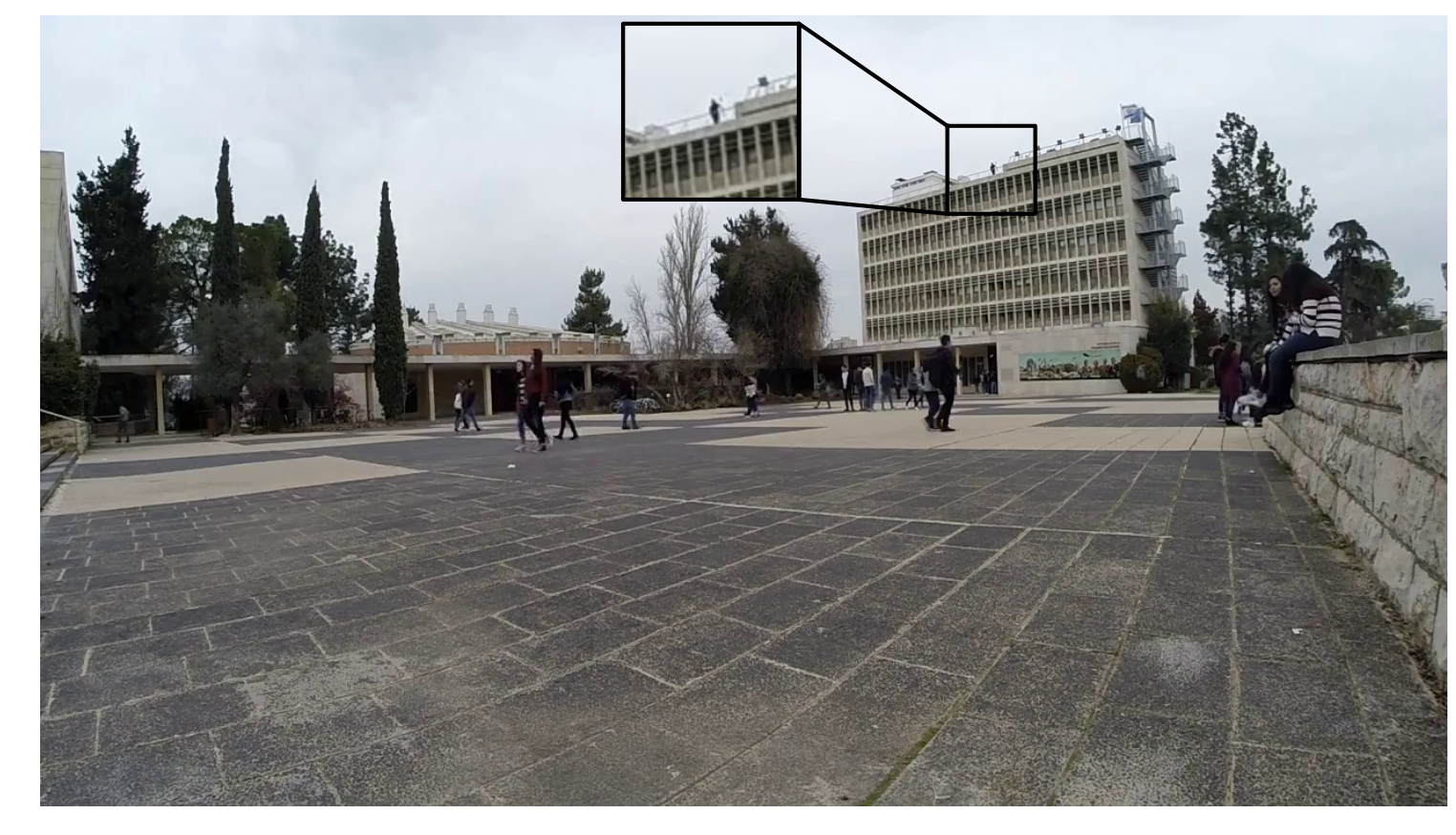}} \\
(a) & (b)
\end{tabular}

 \caption{An example from the Square sequence. (a) An image from the off-plane camera ($B$), with recovered epipolar lines overlayed. The small box zooms-in on the cameraman of the on-plane camera ($A$). (b) A frame from the on-plane wide angle camera, taken at the same time. The area around the other camera is enlarged for  convenience.}
\label{fig:square_closeup}
\end{figure*}

Figure ~\ref{fig:square_closeup} shows an example from a real video with planar motion. A wide angle camera $A$ (GoPro Hero 3+) is mounted at a height of about a meter above ground and facing towards a busy square (right image). Another camera, $B$, captured the same scene from a typical surveillance angle from a nearby roof (left image). 

\begin{figure*}[t]
    \begin{tabular}{cc}
{\includegraphics[width=1\columnwidth]{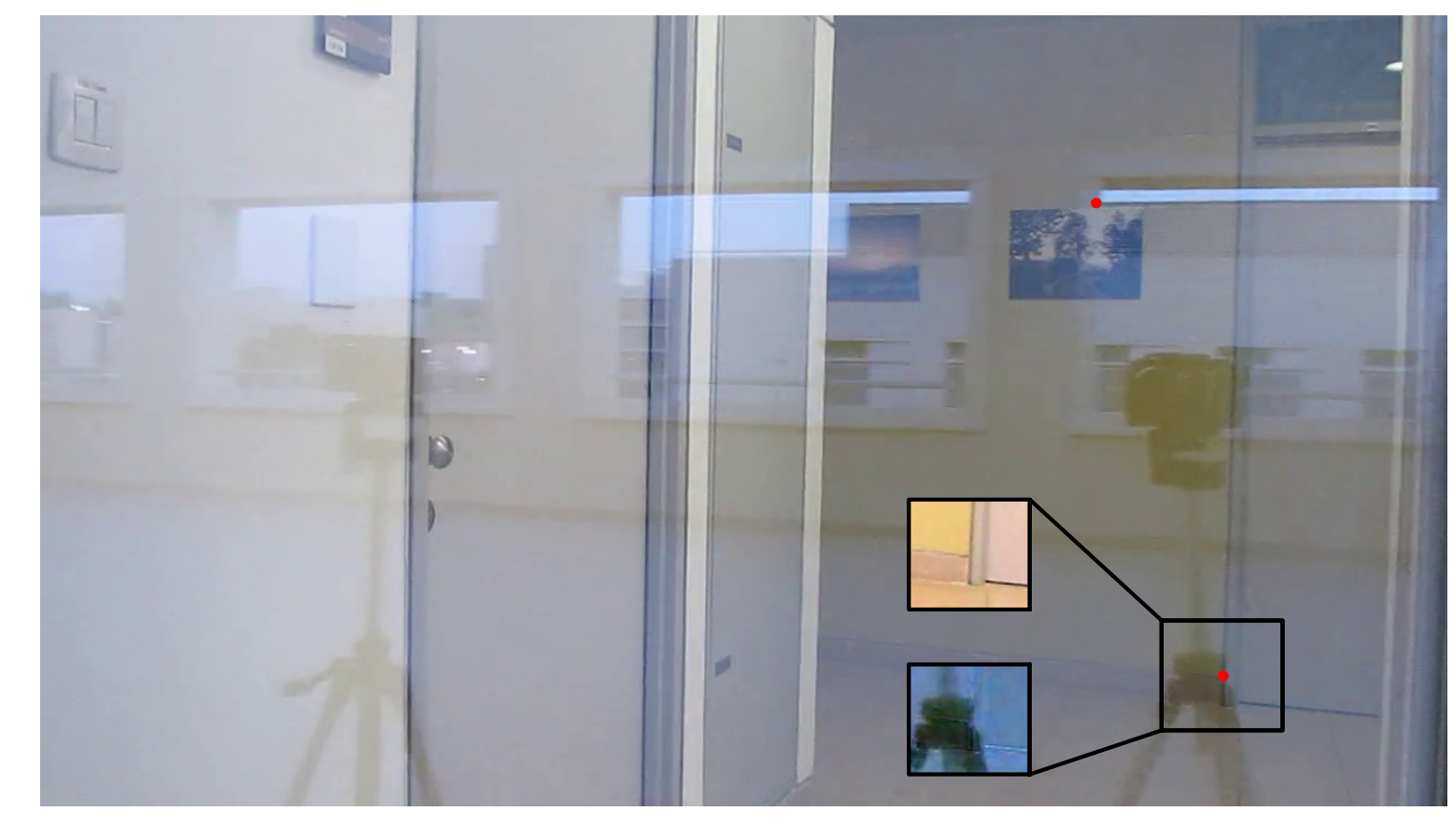}} 
&
{\includegraphics[width=1\columnwidth]{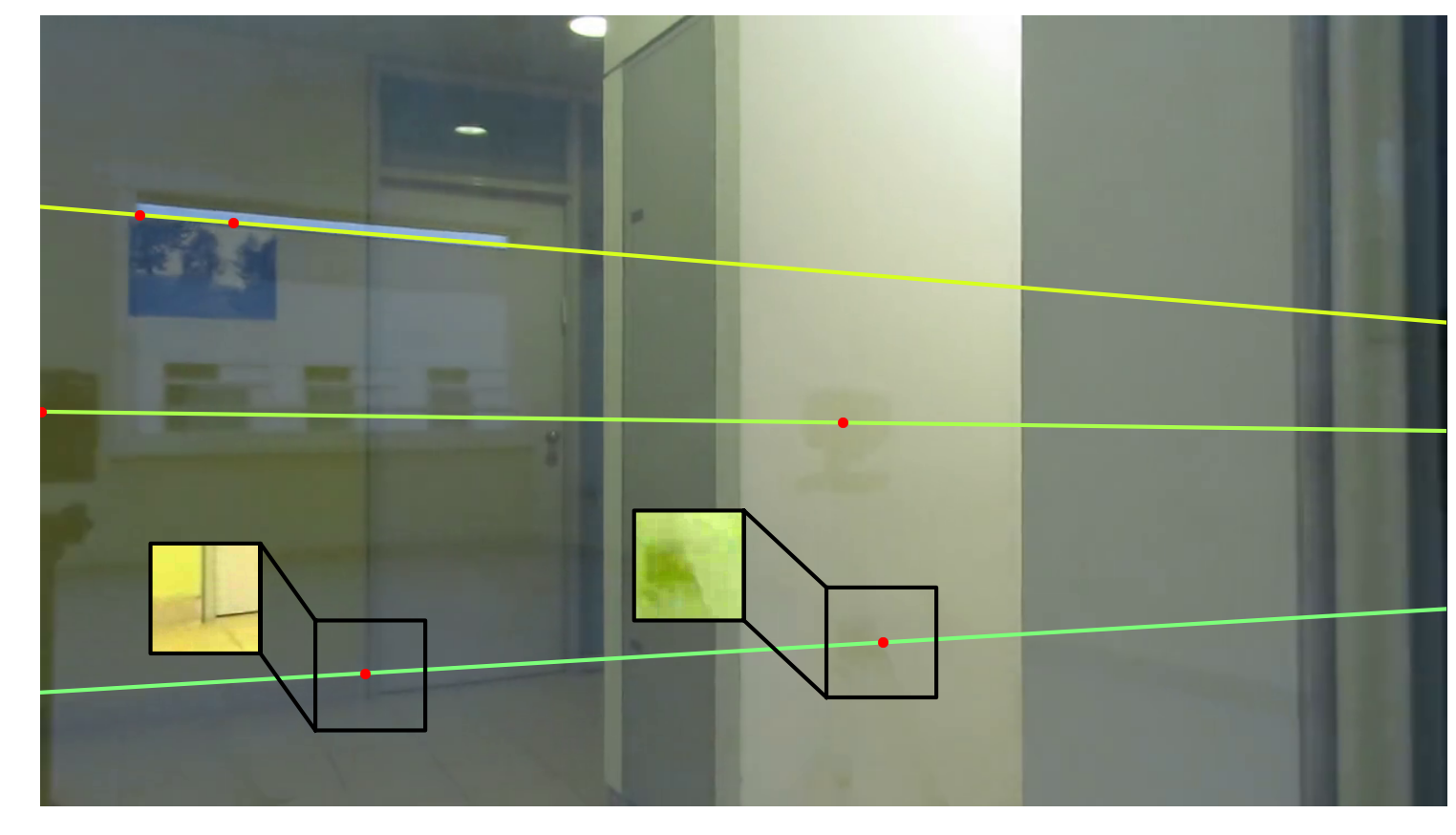}} 
\\
(a) & (b)
\end{tabular}

\caption{In still images of semi transparent surfaces such as windows, multiple objects may be visible at the same image location. (a) Separating  the reflections from the transmitted light  results in two images (highlighted black boxes),  features extracted from these images will correspond to (different) points on an epipolar line in the right image. (b) The two corresponding epipolar lines are shown, as well as a third one, namely the line connecting the reflections of both camera centers.}
\label{fig:glass}
\end{figure*}

An example of images of a static scene with a semi transparent surface is shown in Figure ~\ref{fig:glass}. Behind the flat window, part of a corridor with two doors and a painting on the wall is visible. The reflection on the glass consists of the two cameras with tripods, and the buildings behind. The difference in colors  between the cameras is due to different white balance. The two red dots marked on the left image ($A$) are points where two corner points were detected on different surfaces (one behind the glass and one reflected on it), and the two layers have been separated and shown individually. The two black boxes show a detected corner point on a door and a point on the tripod of camera $A$. The same points are marked with red dots on the right image ($B$). Since the reflecting surface is flat, the virtual location of the reflected tripod is the same for $A$ and $B$. Thus, its projection on $B$ must lie on the same epipolar line as the corner of the door. A second line is obtained by applying the same to the second marked point in $A$, and their intersection yields the epipole. For visualization, the reflections of the two camera centers, which of course share an epipolar line, have also been marked and connected by a line.

\begin{figure*}[t]
	\centering
{\includegraphics[width=1\columnwidth]{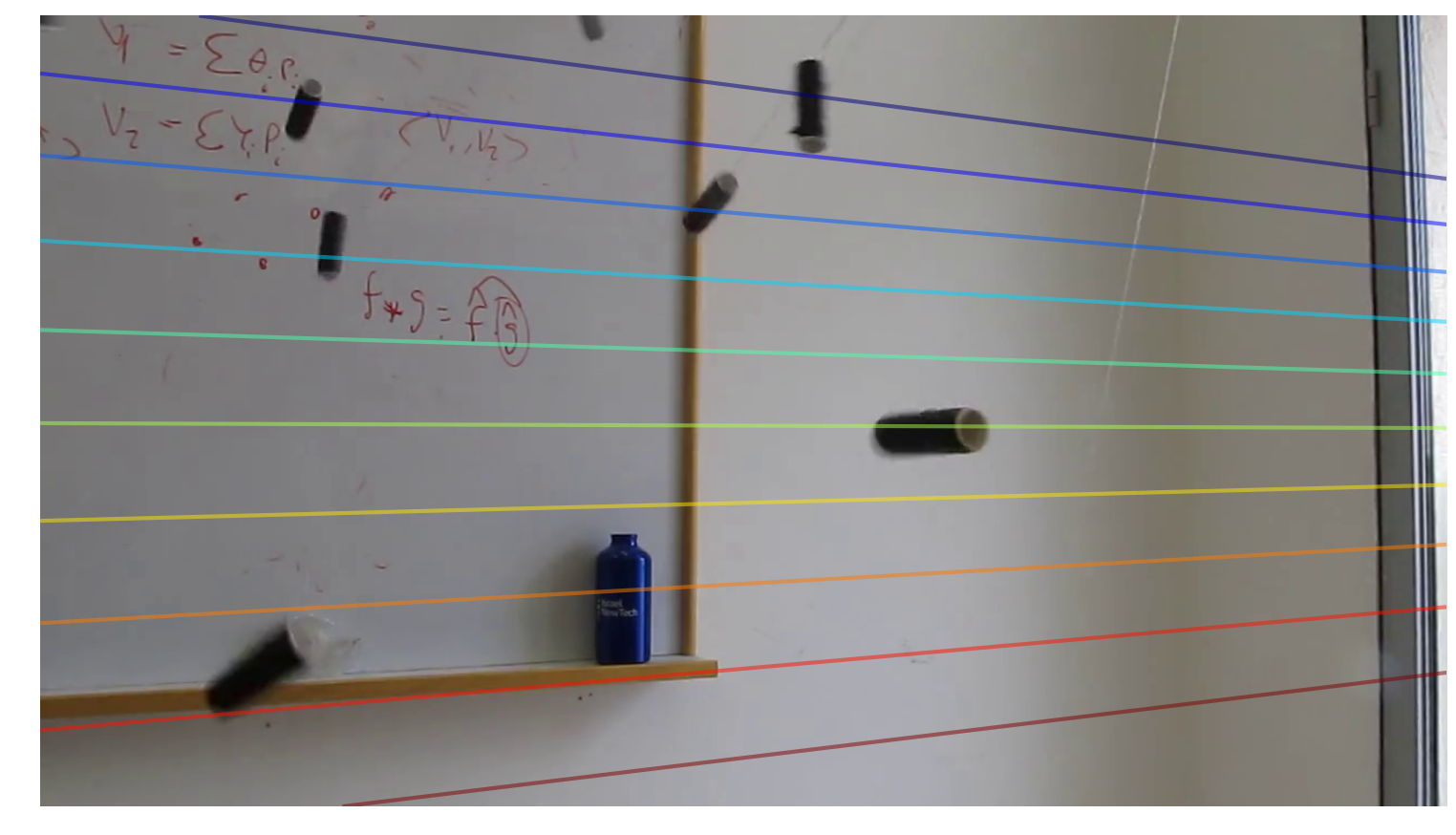}} 
{\includegraphics[width=1\columnwidth]{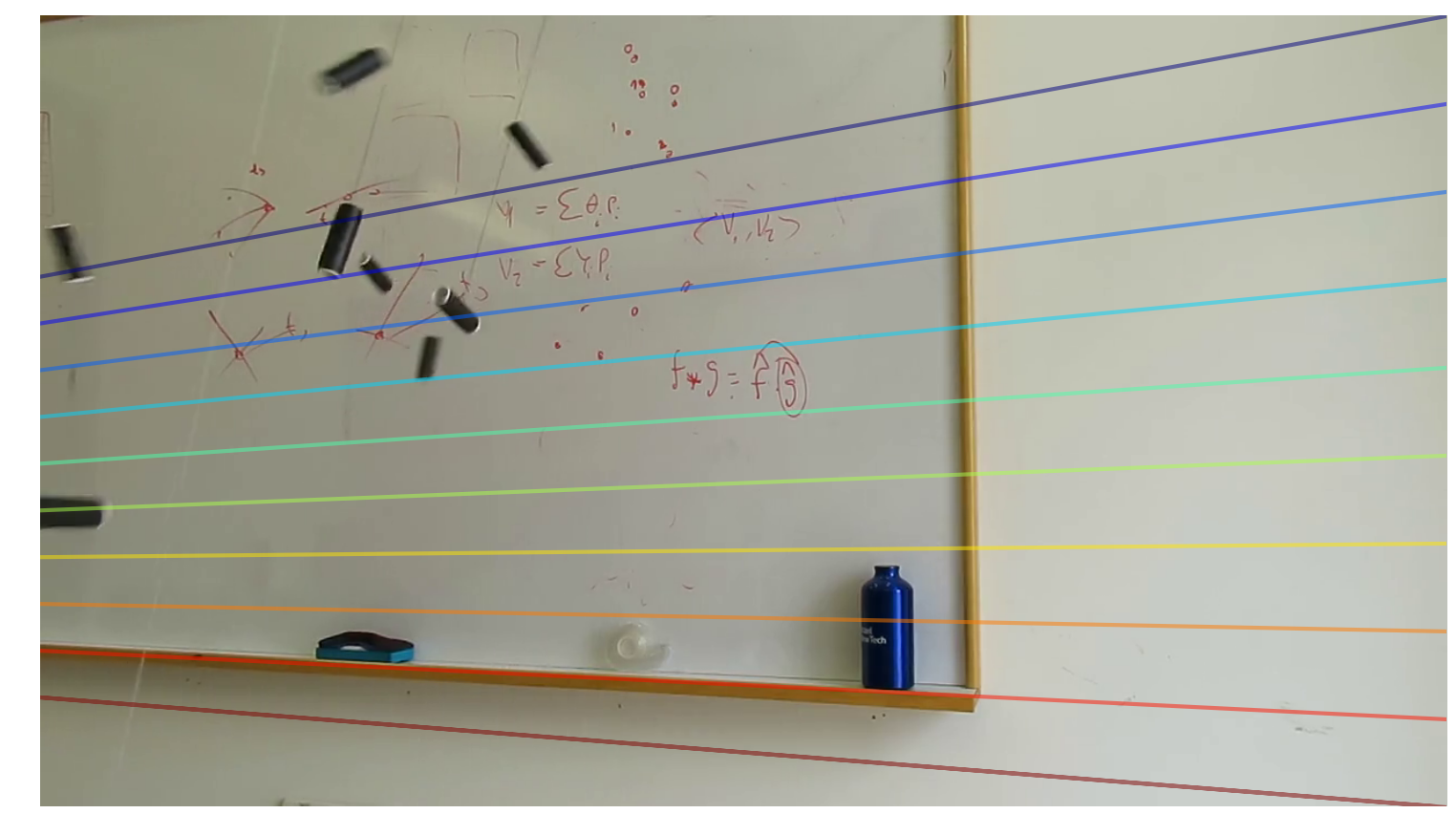}} 

 \caption{A pair of representative frames from the Threads sequence. Recovered pairs of epipolar lines share the same color. Note that although part of the background is visible in both videos, the epipoles cannot be recovered using only corresponding points from the background, since it's essentially planar.}
\label{fig:threads2}
\end{figure*}

Representative samples from other real video experiments are shown in Figures ~\ref{fig:threads2}, ~\ref{fig:fish2}, ~\ref{fig:balls5}, and ~\ref{fig:drone2}. The original videos are given in the supplementary material.

\section{Conclusion}
We introduced a method for finding corresponding epipolar lines
from multiple pixel correspondences in one camera to a single pixel in the other. We conducted experiments with real and synthetic videos, where our method was shown to calibrate cameras with state of the art accuracy while consuming far less computation time, compared to existing methods on a standard dataset.

\section*{Acknowledgements}
The authors would like to thank Or Sharir and Uri Karniel for their help in conducting the experiments.

\clearpage

\begin{figure*}[t]
	\centering
{\includegraphics[width=1\columnwidth]{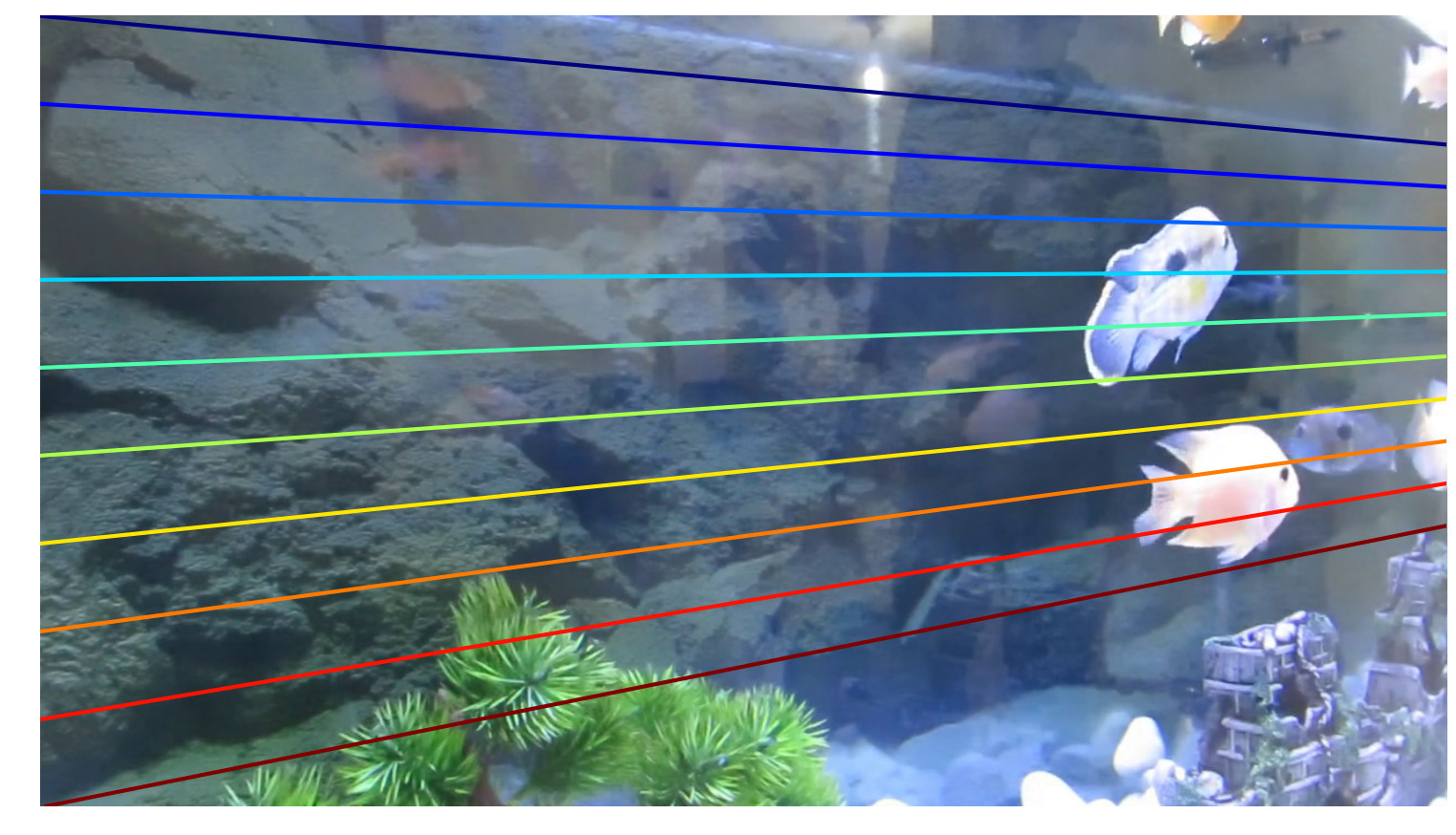}} 
{\includegraphics[width=1\columnwidth]{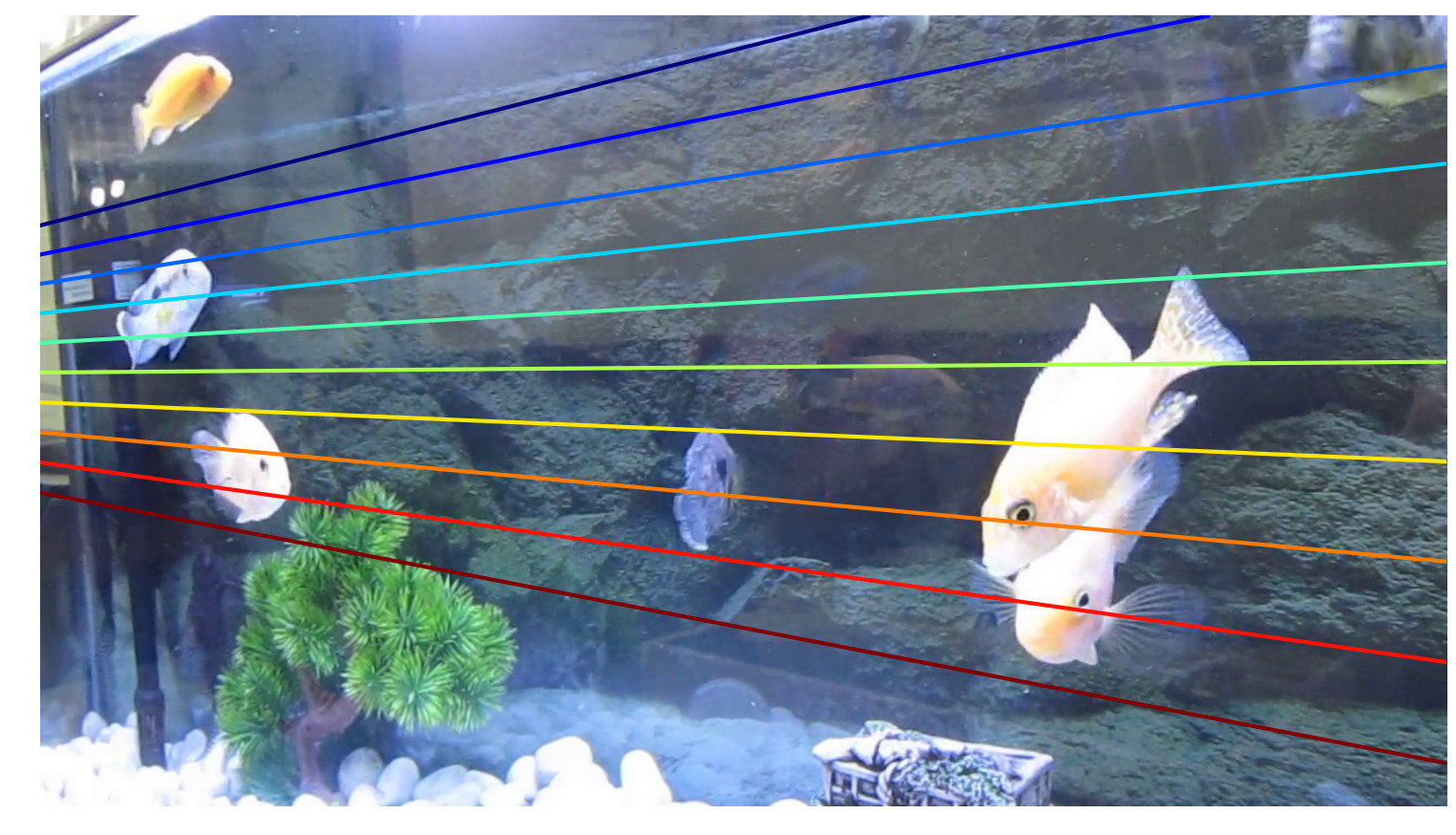}} 

\caption{A pair of representative frames from the Fish sequence with overlaying corresponding epipolar lines. Notice that the camera visible in each of the images is not the other camera, but its reflection on the aquarium wall. The two camera reflections are located on corresponding epipolar lines (turquoise). Best viewed in color.}
\label{fig:fish2}
\end{figure*}

\begin{figure*}[t]
	\centering
{\includegraphics[width=1\columnwidth]{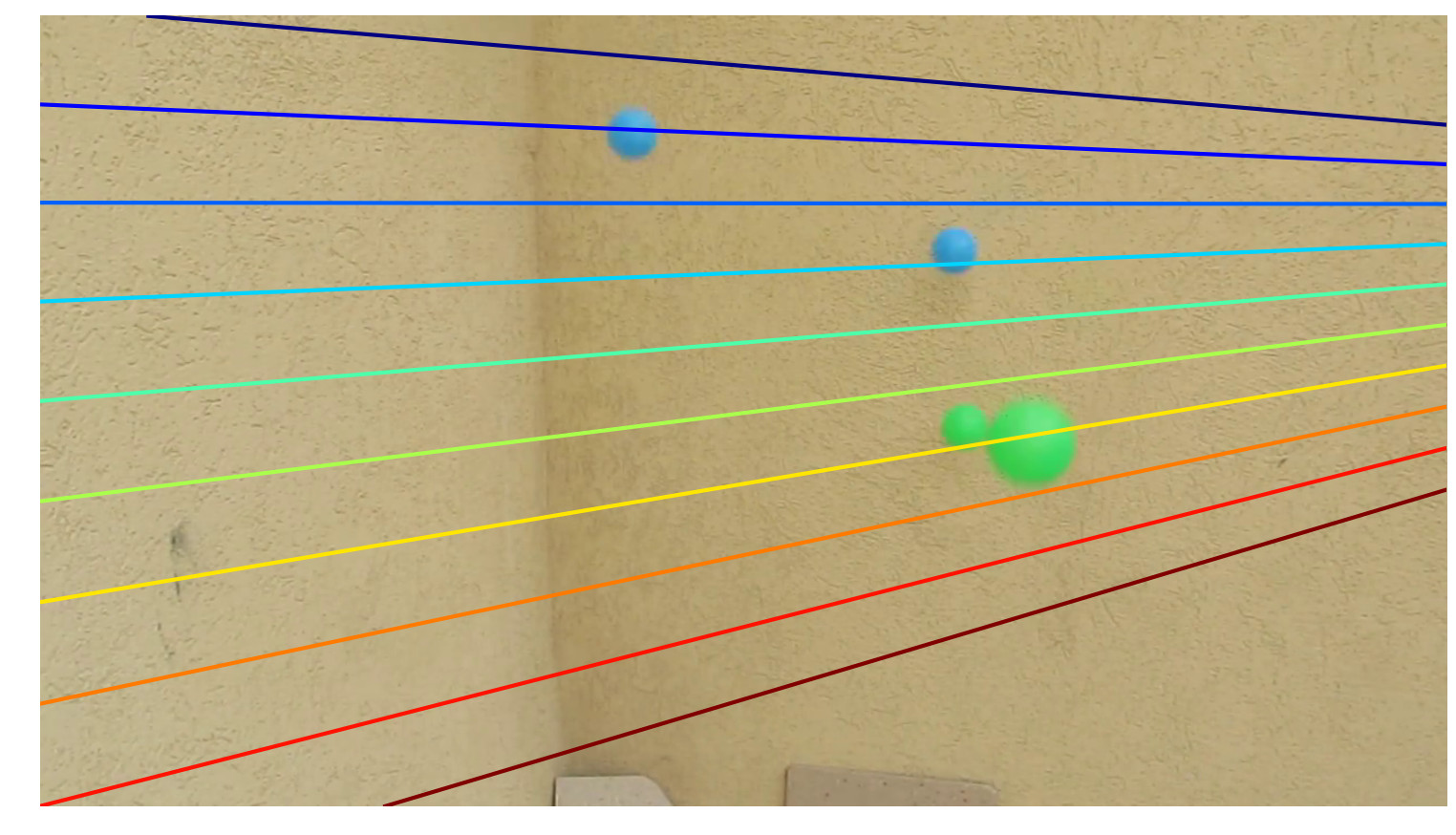}} 
{\includegraphics[width=1\columnwidth]{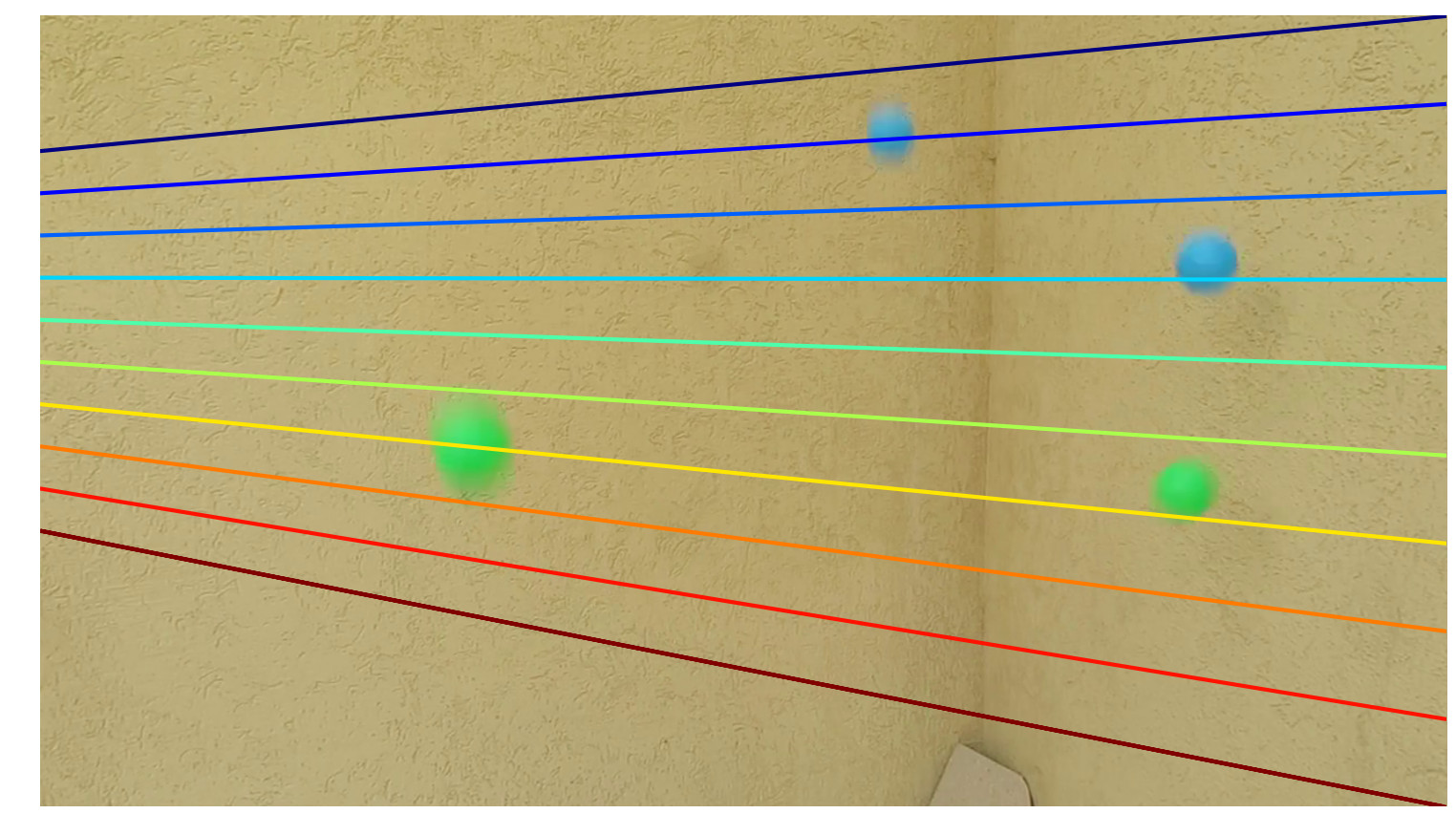}} 

\caption{A representative pair of frames from the Balls sequence. When an  object is a perfect sphere its 3D centroid projects exactly to the center-of-mass of the detected silhouette (up to the precision of the foreground detection).}
\label{fig:balls5}
\end{figure*}

\begin{figure*}[t]
	\centering
{\includegraphics[width=1\columnwidth]{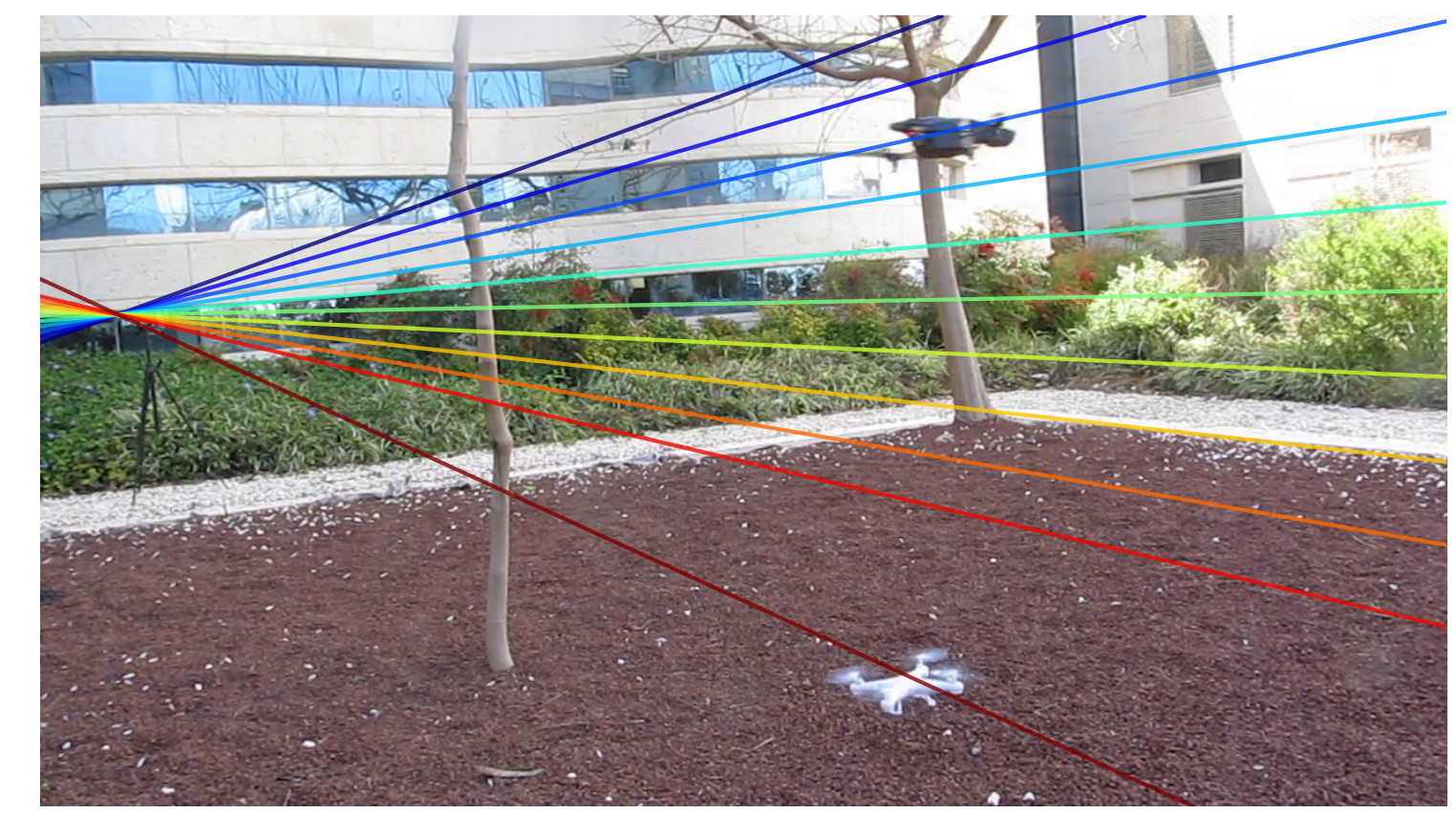}} 
{\includegraphics[width=1\columnwidth]{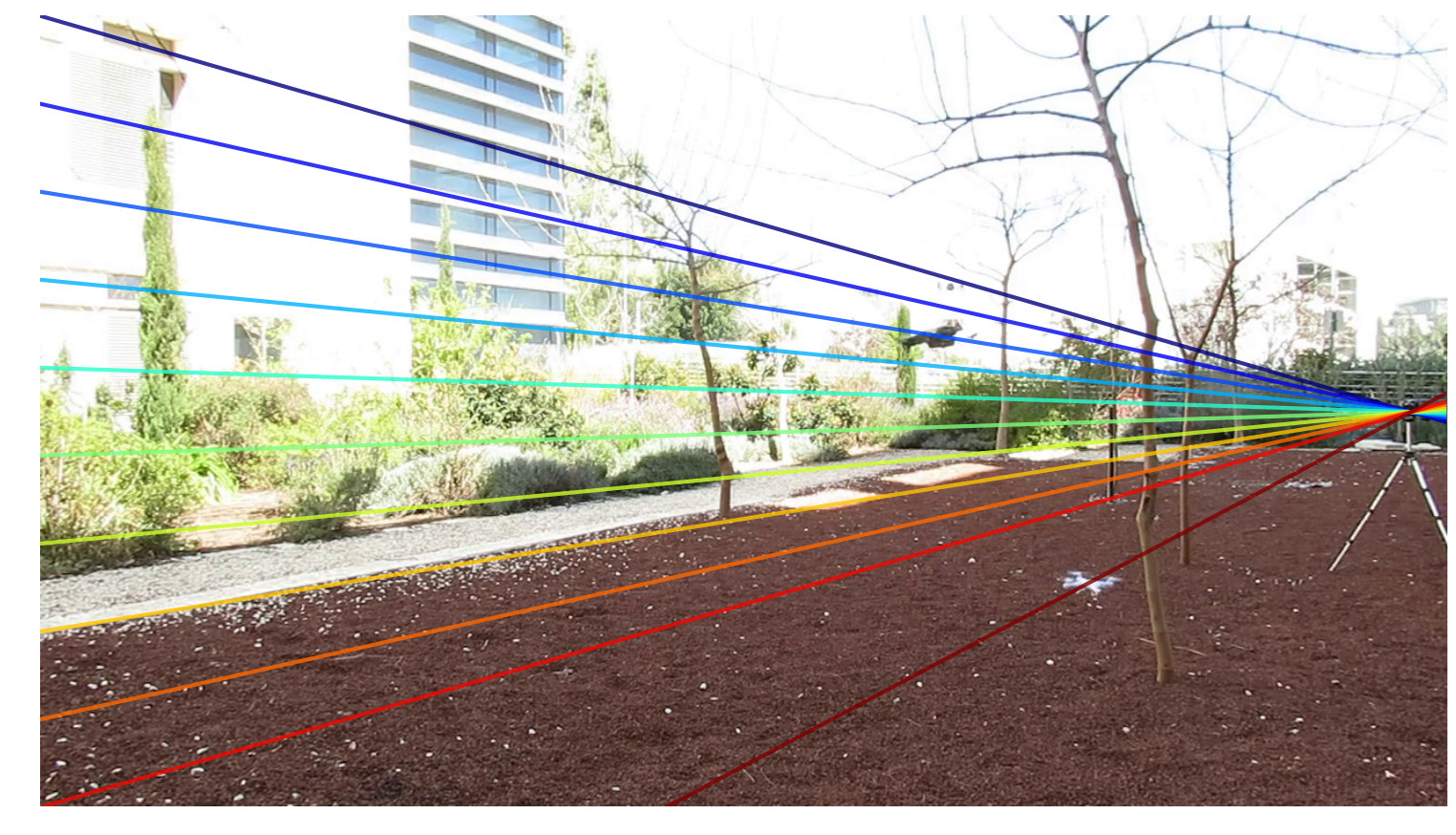}} 

\caption{A representative pair of frames from the Drones sequence. Each camera is visible in the other's field of view.}
\label{fig:drone2}
\end{figure*}

\clearpage

{\small
\bibliographystyle{ieee}
\bibliography{egbib}
}

\end{document}